\documentclass[letterpaper, 10 pt, conference]{ieeeconf}  % Comment this line out if you need a4paper

\usepackage{graphicx}
\usepackage{amsmath,amssymb,amsfonts}
\usepackage{multirow}
\usepackage{algorithm}
\usepackage{algorithmic}
\usepackage{psfrag}
\usepackage{subfigure}
\usepackage{color}
\usepackage{soul}
\usepackage{xcolor}
\usepackage[bookmarks=false]{hyperref}
\usepackage{lipsum}
\usepackage{mathrsfs}
\usepackage{pbox}
\usepackage{flushend}
\usepackage{mathtools}

\makeatletter
\def\hlinewd#1{%
  \noalign{\ifnum0=`}\fi\hrule \@height #1 \futurelet
   \reserved@a\@xhline}
\makeatother

\graphicspath{{Figures/}}

\IEEEoverridecommandlockouts                              % This command is only needed if
\title{\LARGE \bf
Traffic Volume Prediction using Memory-Based Recurrent Neural Networks: A comparative analysis of LSTM and GRU
}

\author{Lokesh Chandra Das
\thanks{Lokesh Chandra Das is with the Department of Computer Science, University of Memphis, Memphis, TN, United States
        {\tt\small \{ldas\}@memphis.edu}}%
}

\begin{document}

\maketitle
\thispagestyle{empty}
\pagestyle{empty}

%%%%%%%%%%%%%%%%%%%%%%%%%%%%%%%%%%%%%%%%%%%%%%%%%%%%%%%%%%%%%%%%%%%%%%%%%%%%%%%%
\begin{abstract}
Predicting traffic volume in real-time can improve both traffic flow and road safety. A precise traffic volume forecast helps alert drivers to the flow of traffic along their preferred routes, preventing potential deadlock situations. Existing parametric models cannot reliably forecast traffic volume in dynamic and complex traffic conditions. Therefore, in order to evaluate and forecast the traffic volume for every given time step in a real-time manner, we develop non-linear memory-based deep neural network models.  Our extensive experiments run on the Metro Interstate Traffic Volume dataset demonstrate the effectiveness of the proposed models in predicting traffic volume in highly dynamic and heterogeneous traffic environments. 
\end{abstract}

%%%%%%%%%%%%%%%%%%%%%%%%%%%%%%%%%%%%%%%%%%%%%%%%%%%%%%%%%%%%%%%%%%%%%%%%%%%%%%%%

\section{Introduction}
\label{sec:introduction}
Rapid socioeconomic development aids in the growth of large-scale, expanding smart cities with easy access to communication technologies. Modern mobile and vehicular communication technology promotes Intelligent Transportation Systems (ITS), resulting in an exponential increase in the number of vehicles on the road each year. Traffic congestion has become one of the most pressing issues to address\cite{fan2013multi} and it is creating deadlock situations for large cities as well as for medium and small cities\cite{kuang2020traffic}. Traffic congestion can be mitigated using traditional methods by changing the road and urban infrastructures. However, redesigning the city structure to improve the traffic pattern and reduce congestion is very expensive, let alone time-consuming. Therefore, dynamic route planning, optimizing road allocations, managing traffic on urban roads, and using modern technologies to understand traffic patterns better are crucial tasks to reduce traffic congestion successfully. Forecasting the future traffic status based on past historical data is a way of reducing traffic congestion\cite{yu2017deep}. An accurate traffic volume prediction can alleviate traffic congestion and optimize traffic distributions.\\
\newline
Different techniques for traffic flow prediction have been studied lately. These methods can be generally classified as naive, parametric, and non-parametric. The naive model does not make any assumptions, and it is computationally fast. However, it has low accuracy. The parametric method includes different time-series methods. One of the most widely used parametric methods is the ARIMA (autoregressive integrated moving average) model\cite{chatfield2000time}. The ARIMA model is practically effective in predicting traffic flow and has been a benchmark. However, the parametric approaches can achieve better performances if the time-series data shows a regular pattern. In particular, the parametric methods fail to perform better when the traffic pattern varies in nature. Non-parametric methods address the issue of parametric methods, and various non-parametric methods such as non-parametric regression, support vector machine, Kalman filtering, and neural network predictors are being used in traffic volume predictions. Deep neural networks have been shown to be superior at dealing with traffic forecasting. Recurrent neural networks (RNN), especially long-short Term memory (LSTM), demonstrated their advantages in modeling and predicting traffic flow.\\

In this project, we develop traffic volume prediction using long-short term memory (LSTM) and gated recurrent units (GRU) and evaluate our model on Metro inter-state traffic dataset\cite{hogue2019metro}.

\section{Related Work}
\label{sec:related_work}
Traffic forecasting is critically important, and it has been extensively studied lately. Traffic prediction methods can be divided into parametric and non-parametric methods. Previously, researchers have applied various parametric methods to predict traffic flow. The autoregressive integrated moving average is one of the most widely used parametric models in time-series prediction datasets. Chen et al.\cite{chen2011short} predicted traffic flow using the autoregressive integrated moving average (ARIMA) model. However, other parametric models have also been used by researchers. Kumar et al.\cite{kumar2017traffic} developed Kalman filtering techniques to forecast the traffic flow. Dong et al.\cite{dong2018short} applied the gradient-boosting decision tree algorithm to perform short-term traffic flow prediction. \\\\
Deep learning-based approaches have recently received a lot of attention from researchers, and they provide a more accurate estimation of traffic volume prediction than traditional parametric methods. The long-short-term memory (LSTM) and gated recurrent unit (GRU) is capable of holding a long sequence of past observations and making a correlation in sequence prediction, which makes them more suitable for time-series datasets. Many researchers are using LSTM and GRU for predicting traffic conditions. Zhao et al.\cite{zhao2017lstm} used LSTM to predict short-time traffic. Fu et al.\cite{fu2016using} used LSTM and GRU for traffic flow prediction. Some researchers use a graph attention network to predict the traffic pattern\cite{zheng2020gman}. In this project, we use LSTM and GRU to predict future traffic flow and evaluate the model using a metro interstate traffic volume dataset.

\subsection{Problem Formulation}
\label{sec:problem_formulation}
We consider the real-time traffic volume prediction as a multi-variate time-series problem where our model will approximately estimate future traffic flow based on the current and $t-$hours of historical observations. Specifically, our objective is to foretell future traffic volume at time step $T_{t+1}, T_{t+2},...T_{t+f}$, where $f$ is the future prediction horizon using past observations from time steps $T_{t-l}, T_{t-l-1}, ..., T_{t-1}, T_t$ where $l$ is the length of past observations used to predict the future traffic flow.
\section{Method}
We use memory-based recurrent neural network models, e.g., long-short-term memory (LSTM) and gated recurrent units (GRU), to predict future traffic flow.
\subsection{Long-Short Term Memory (LSTM)}

\begin{figure}[ht!]
     \centering
     \begin{subfigure}
         \centering
         \includegraphics[width=0.4\textwidth]{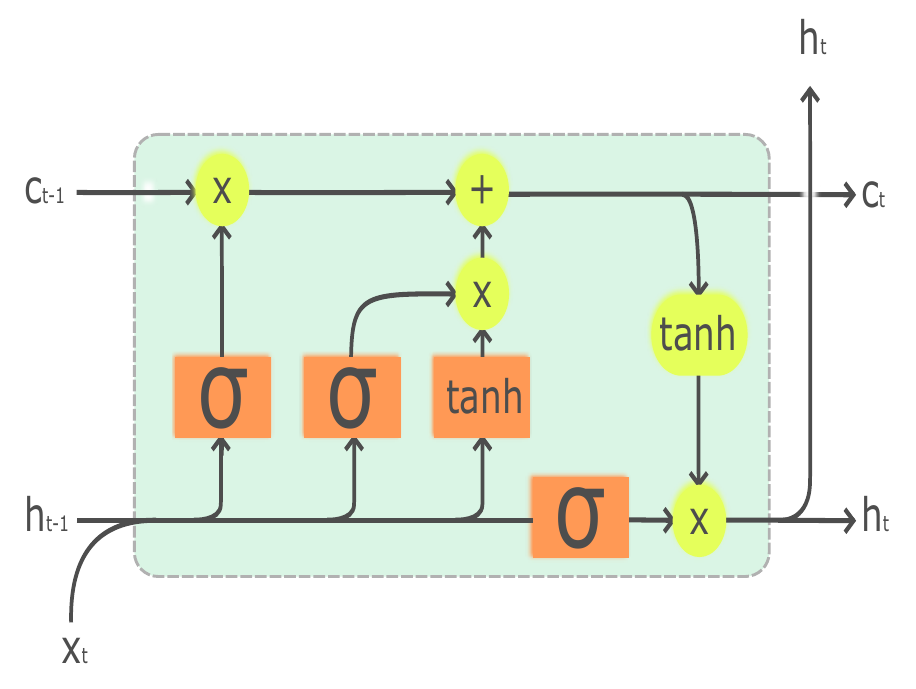}
         \caption{A structure of an LSTM cell}
         \label{fig:lstm}
     \end{subfigure}%
     \begin{subfigure}
         \centering
         \includegraphics[width=0.4\textwidth]{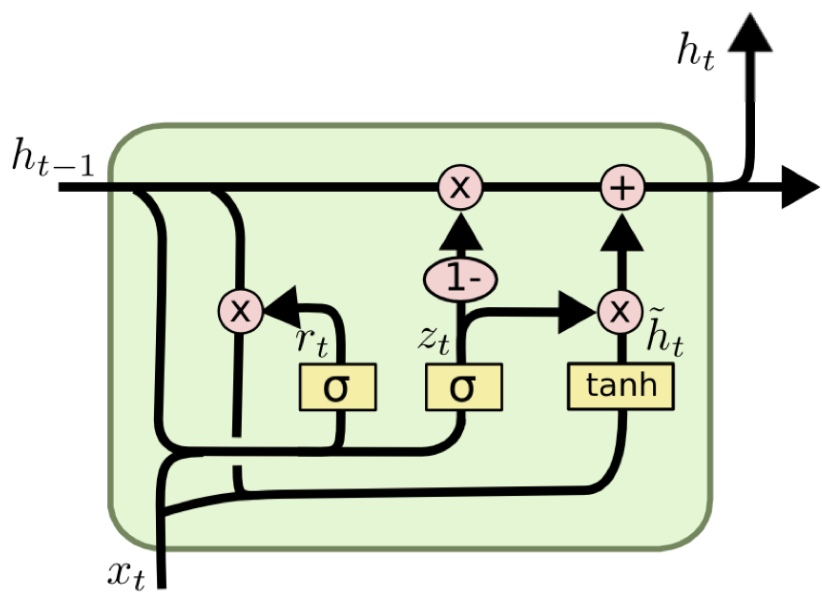}
         \caption{A structure of a GRU cell}
         \label{fig:gru}
     \end{subfigure}
\end{figure}

Long-Short-term Memory networks (LSTMs)\cite{hochreiter1997long} are designed to learn long-term dependencies and effectively deal with the vanishing gradient problem of recurrent neural networks(RNNs)\cite{li2019lstm,hou2019normalization}. Because it can hold long sequences while predicting the current output, LSTM is well-suited for applications that make predictions based on time-series data. The memory block of the LSTM cell makes it easy to hold the sequence information. The memory block has memory cells and three gates: the forget gate, the input gate, and the output gate. The basic structure of an LSTM cell is depicted in Fig.\ref{fig:lstm}.
% inset LSTM figure

The forget gate determines which information should propagate for the next time sequence through the sigmoid activation function. The input gate decides which information is necessary for the current state, and the output gate regulates what to output to the next state, e.g., the current output and the value of the next hidden state $h_t$. The equations to update the gates are as follows:
\begin{equation*}
    \label{lstm_quation}
    \begin{split}
        f_t &= \sigma(W_f.[h_{t-1}, x_t]+b_f)\\
        i_t &= \sigma(W_i.[h_{t-1}, x_t]+b_i)\\
        \Tilde{C_t} &= \tanh(W_C.[h_{t-1}, x_t]+b_C)\\
        C_t &= f_t*C_{t-1}+i_t*\Tilde{C_t}\\
        o_t &= \sigma(W_o.[h_{t-1}, x_t]+b_o)\\
        h_t &= o_t*\tanh(C_t)
    \end{split}
\end{equation*}
where $f_t$ is the forget gate, $i_t$ is the input gate, and $o_t$ is the output gate, respectively; $C_t$ is a memory cell to hold the sequences from the previous states, and $h_t$ is the output for the next state; $W^*$ is weight, and $b^*$ is bias.
\subsection{Gated Recurrent Unit (GRU)}
Gated Recurrent Unit (GRU) replaces the three gates of LSTM with two gates: the reset gate and the update gate. These gates use sigmoid activation functions linked by LSTMs, constraining their values between $0$ and $1$. Intuitively, the reset gate controls how much of the previous state we might still want to remember, and an update gate would allow us to control how much of the new state is just a copy of the old state. Fig. \ref{fig:gru} illustrates the inputs for both the reset and update gates in a GRU, given the input of the current time step and the hidden state of the previous time step. The mathematical mechanism for the GRU is as follows:
\begin{equation*}
    \label{gru_quation}
    \begin{split}
        r_t &= \sigma(W_r.[h_{t-1}, x_t]+b_r)\\
        z_t &= \sigma(W_z.[h_{t-1}, x_t]+b_z)\\
        \Tilde{H_t} &= \tanh(W_h.[(h_{t-1}\odot r_t), x_t]+b_h)\\
        h_t & = z_t\odot h_{t-1} + (1-z_t)\odot \Tilde{H_t}
    \end{split}
\end{equation*}
where $r_t$ is the reset gate, $z_t$ is the update gate, and $h_t$ is the current time step's output. The rest of the symbols contain the same meaning as described in the LSTM section.
\subsection{Dataset}
We use Metro Interstate Traffic Volume Data Set\cite{hogue2019metro}. This is a multivariate, sequential, time-series dataset collected
from the westbound I-94 at Minneapolis-St. Paul, Minnesota. The dataset is relatively large and has 48204 instances with nine features. The dataset collected traffic volume data in an hourly manner from 2012 to 2018, taking into account weather features and holidays that impact the traffic volume.
\subsection{Data Preprocessing}
\begin{figure}[ht!]
    \centering
    \includegraphics[width=0.5\textwidth]{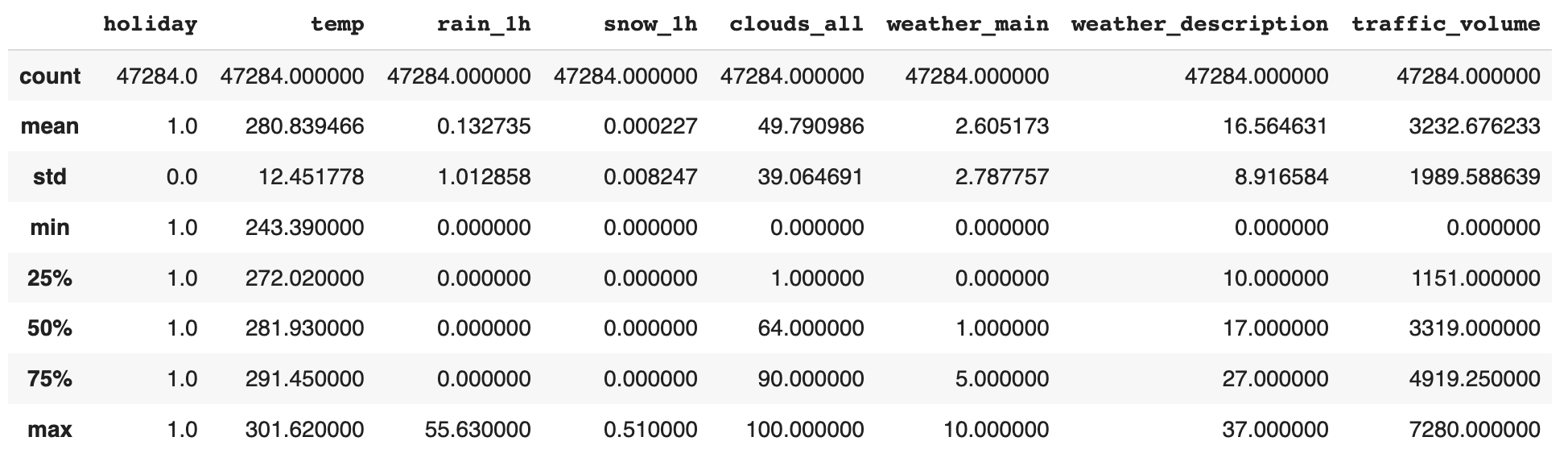}
    \caption{Statistics of the dataset}
    \label{fig:data_summary}
\end{figure}

\begin{figure}[ht!]
     \centering
     \begin{subfigure}
         \centering
         \includegraphics[width=0.5\textwidth]{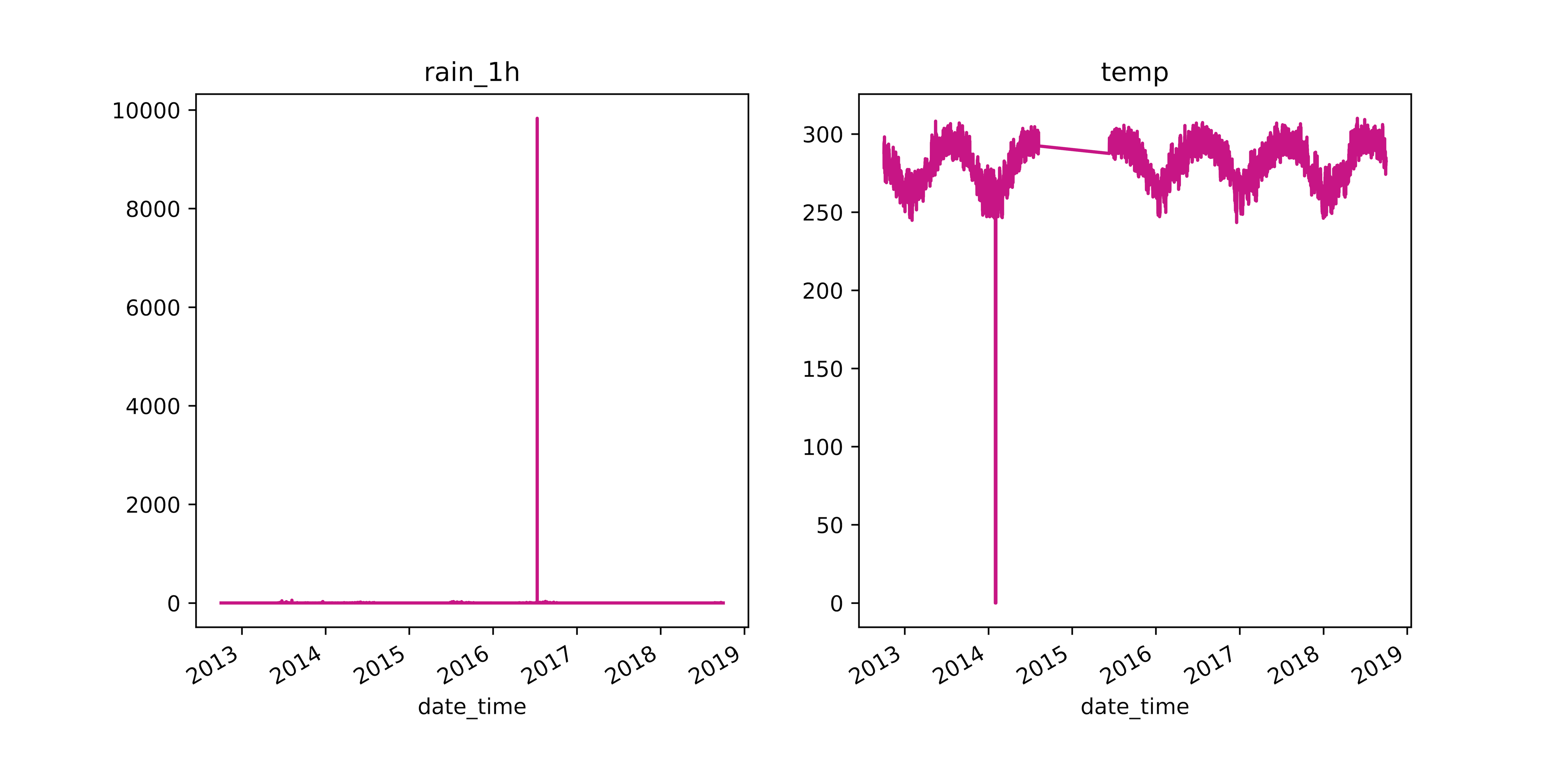}
         \caption{Features contains outliers}
         \label{fig:outlier}
     \end{subfigure}
     \begin{subfigure}
         \centering
         \includegraphics[width=0.5\textwidth]{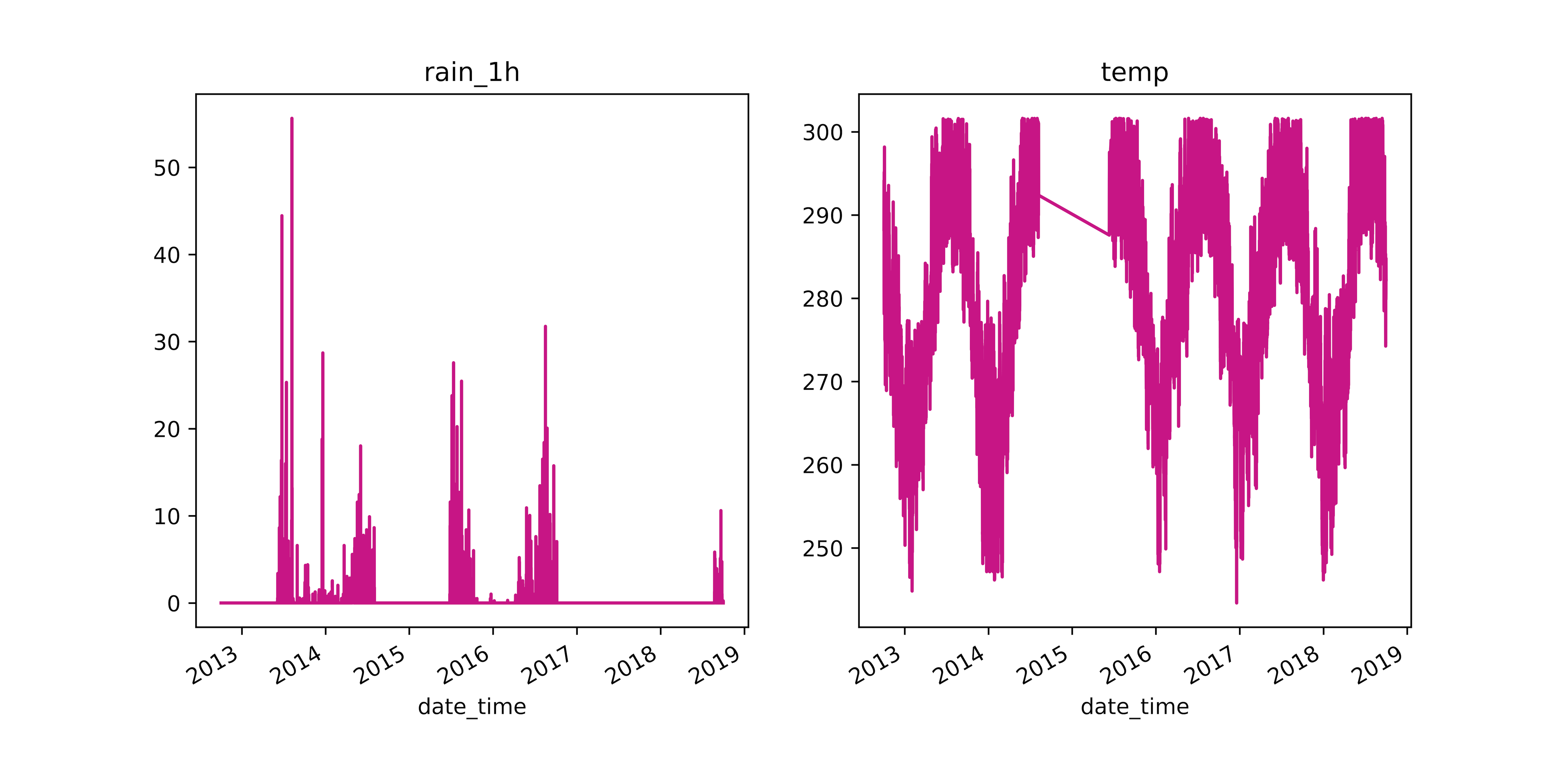}
         \caption{Features after removing outliers}
         \label{fig:without_outlier}
     \end{subfigure}
     %\caption{Dataset contains outliers}
     \label{fig:outlier_remover}
\end{figure}

We cannot directly feed data into the LSTM or GRU models, unlike other deep learning models such as CNN and RNN. We have to convert it into a specific format. Input format should have at least time steps and a number of features. Generally, in time series prediction, we use $t$-hours of observations as input to the network, and the model will produce output at $t+1$ hour. Here, we use the past $t$-hours of data to predict the next $n$ hours' traffic volume. The dataset also contains some categorical values that needed to be converted into numerical values. Moreover, the attributes are in different scales. The statistics of the dataset are shown in Fig.\ref{fig:data_summary}. It is clear that traffic\_volume attribute values are very larger compared to rain\_1h values. So we need to scale them to reduce the bias. We use the MinMaxScaler technique to normalize the feature values between 0 and 1. We split the dataset into training and testing sets. We use data from the years 2012-2017 for training among which 20\% was used for validation purposes and the last year's data was used for testing purposes.
From the extensive data analysis, we found that some features contained outliers. For example, rain\_1h and temp features have outliers. We remove them before feeding them into the model to improve its generalization capability. We use the interquartile range~\cite{vinutha2018detection} technique to remove the outlier. Everything that is 1.5 times less than the first interquartile range or 1.5 times greater than the last interquartile range is removed. Figures in Fig. \ref{fig:outlier} and \ref{fig:without_outlier} respectively show the features before and after applying the interquartile range outlier remover technique.

\section{Experimental Setup}
We run a total of 4 experiments with different settings, taking 6 to 24 hours of past observations to predict the next hour's traffic volume. We also did an experiment to see how the features impacted traffic volume prediction. In one set of experiments, we select all features from the dataset, and in another set of experiments, we only consider four features, namely \textit{temperature}, \textit{rain\_1h}, \textit{clouds\_all}, and \textit{traffic volume}. We run experiments on two sets of neural networks. The table \ref{tab:hyperparameters} shows the hyperparameters for our deep learning model. For experiment 2, we just changed the hidden number of units as like $[256, 128, 64, 32]$ and the number of epochs is $500$. The rest of the hyperparameters are the same as those shown in the table.
    \begin{table}[ht]
    \centering
    \caption{Table}
    \label{tab:hyperparameters}
    \resizebox{0.5\textwidth}{!}{%
    \begin{tabular}{|cc|}
    \hline
    \multicolumn{2}{|c|}{Neural Network Settings} \\ \hline
    \multicolumn{1}{|c|}{Input Shape} & (\# of Records, Time Step, \# of Features) \\ \hline
    \multicolumn{1}{|c|}{Hidden Layers} & \begin{tabular}[c]{@{}c@{}}4 Hidden Layers with 128, 64, 32, 16 \\ units of neurons in each layer respectively\end{tabular} \\ \hline
    \multicolumn{1}{|c|}{Activate Function} & tanh \\ \hline
    \multicolumn{1}{|c|}{Batch Size} & 64 \\ \hline
    \multicolumn{1}{|c|}{Learning Rate} & 0.0001 with decay rate 1e-5 \\ \hline
    \multicolumn{1}{|c|}{Optimizer} & Adam \\ \hline
    \multicolumn{1}{|c|}{Epochs} & 300 \\ \hline
    \end{tabular}%
    }
    \end{table}
The training is stopped if the validation error does not improve for at least five consecutive runs. The main evaluation metrics are mean squared error (MSE), mean absolute error (MAE), and mean absolute percentage error (MAPE) as shown in equations \ref{mse}, \ref{mae}, and \ref{mape}.
\begin{equation}
    \label{mse}
    \begin{split}
        MSE(y, \hat{y}) &= \frac{1}{n}\sum_{i=1}^{n}(y_i-\hat{y_i})^2
    \end{split}
\end{equation}
\begin{equation}
    \label{mae}
    \begin{split}
        MAE(y, \hat{y}) &= \frac{1}{n}\sum_{i=1}^{n}|y_i-\hat{y_i}|
    \end{split}
\end{equation}
\begin{equation}
    \label{mape}
    \begin{split}
        MAPE(y, \hat{y}) &= \frac{1}{n}\sum_{i=1}^{n}\frac{|y_i-\hat{y_i}|}{\text{max}(\epsilon, |y_i|)}
    \end{split}
\end{equation}
where $y$ is actual traffic volume, $y_i$ is the predicted traffic volume and $\epsilon \ll 1$ used to avoid any undefined results caused by if $|y_i|$ is zero.
\section{Results}
We implemented the traffic volume prediction problem in Python based on Keras and Tensorflow. To train and test the proposed model, a workstation equipped with an Intel Xeon Gold 5222 processor, an NVIDIA® RTXTM A4000 graphics card, and 48GB of RAM running Windows 11 OS is used.\\
Empirically, the autoregressive integrated moving average (ARIMA) is not suitable when the dataset is more complex and has long sequences. So, we directly choose LSTM and GRU, and ARIMA is out of the scope of the project. In our experiment, we compare the LSTM and GRU performances by varying the neural network settings and also by varying the length of historical observations. We compare the mean squared error (MSE), mean absolute error (MAE), and mean absolute percentage error (MAPE) for both GRU and LSTM with different neural network settings, taking into account all features vs. a reduced number of features.
\begin{figure}[ht!]
    \centering
    \includegraphics[width=0.5\textwidth]{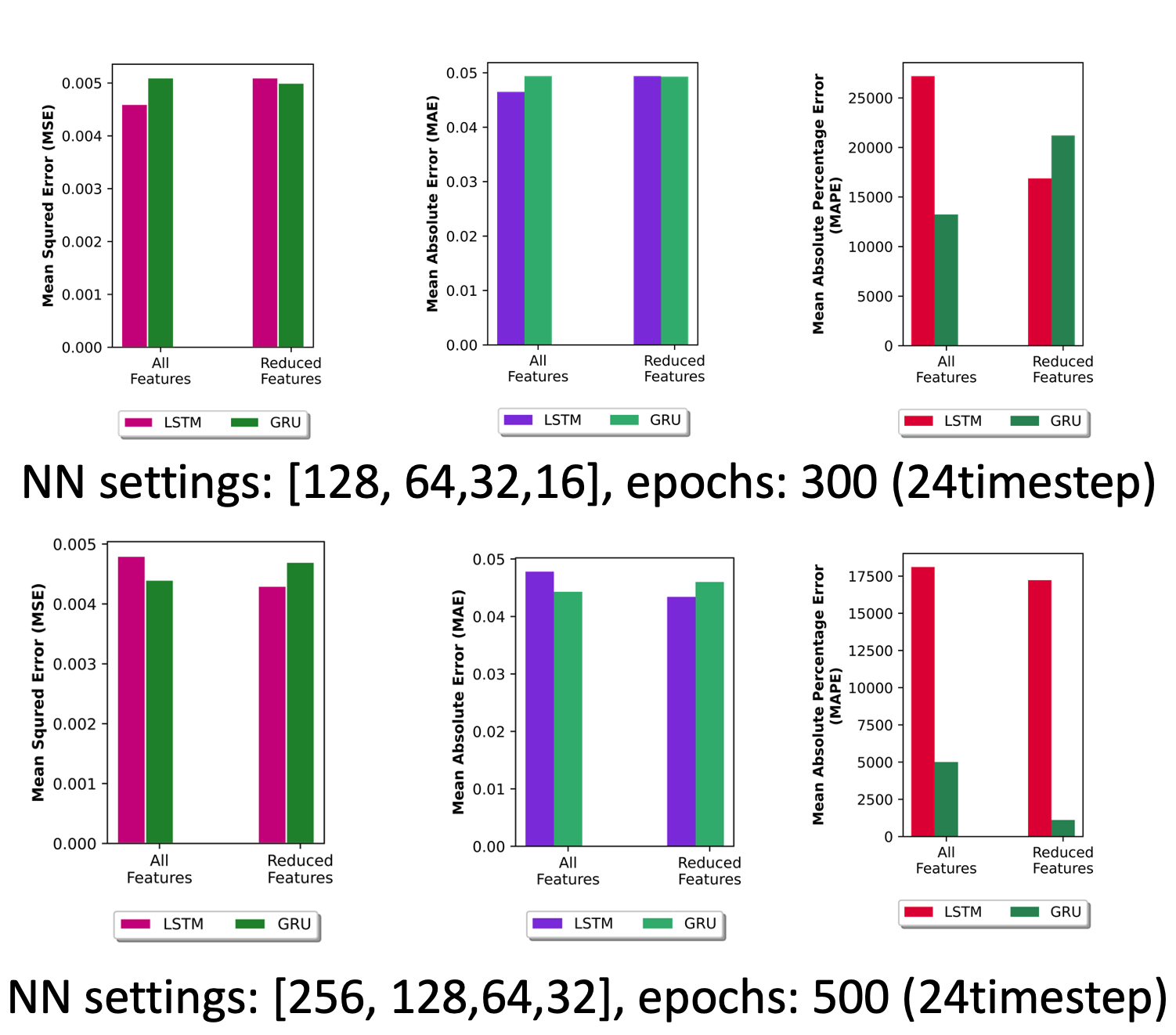}
    \caption{MSE, MAE, and MAPE errors of LSTM and GRU by varying neural network settings. The network uses 24 hours of observations to predict the next hour's traffic volume.}
    \label{fig:eva1}
\end{figure}
Fig. \ref{fig:eva1} shows the comparative results of two models (e.g., LSTM and GRU) for different feature sets and neural network settings when the length of past observation is set to 24 hours. Although there is no significant pattern in the results, it can be seen that both the GRU and LSTM models produce better results in terms of MSE and MAE.
\begin{figure}[ht!]
    \centering
    \includegraphics[width=0.5\textwidth]{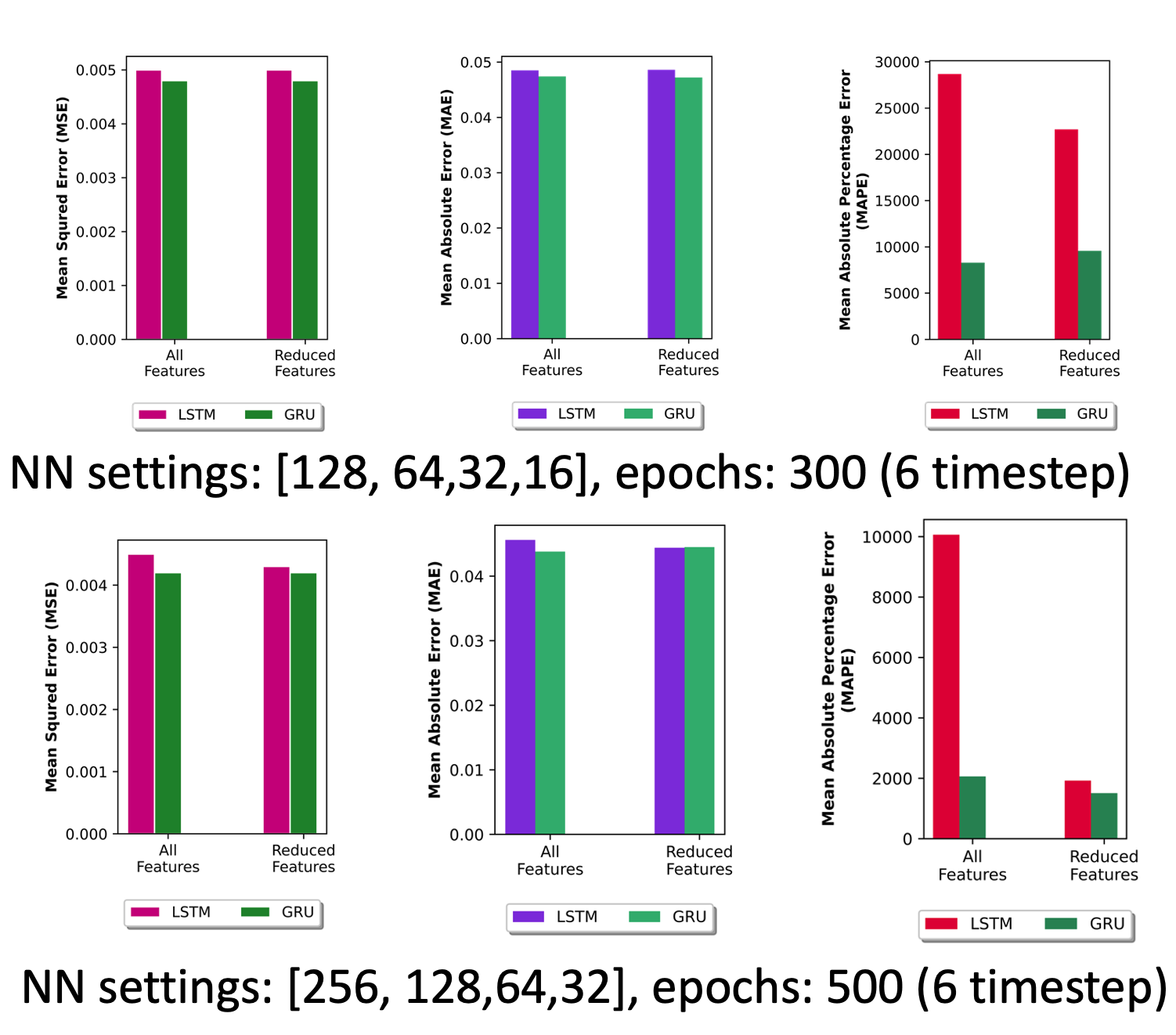}
    \caption{MSE, MAE, and MAPE errors of LSTM and GRU by varying neural network settings. The network uses 6 hours of observations to predict the next hour's traffic volume.}
    \label{fig:eva2}
\end{figure}
On the other hand, when training does not require preserving a very long history, GRU performs better than LSTM. Fig.\ref{fig:eva2} shows the results when the length of the past observation is set to 6 hours. It can be observed that when the past observation history is very limited, GRU performs better compared with the LSTM in all three evaluation metrics for both neural network settings.
\begin{figure*}[ht!]
    \centering
    \includegraphics[width=\textwidth]{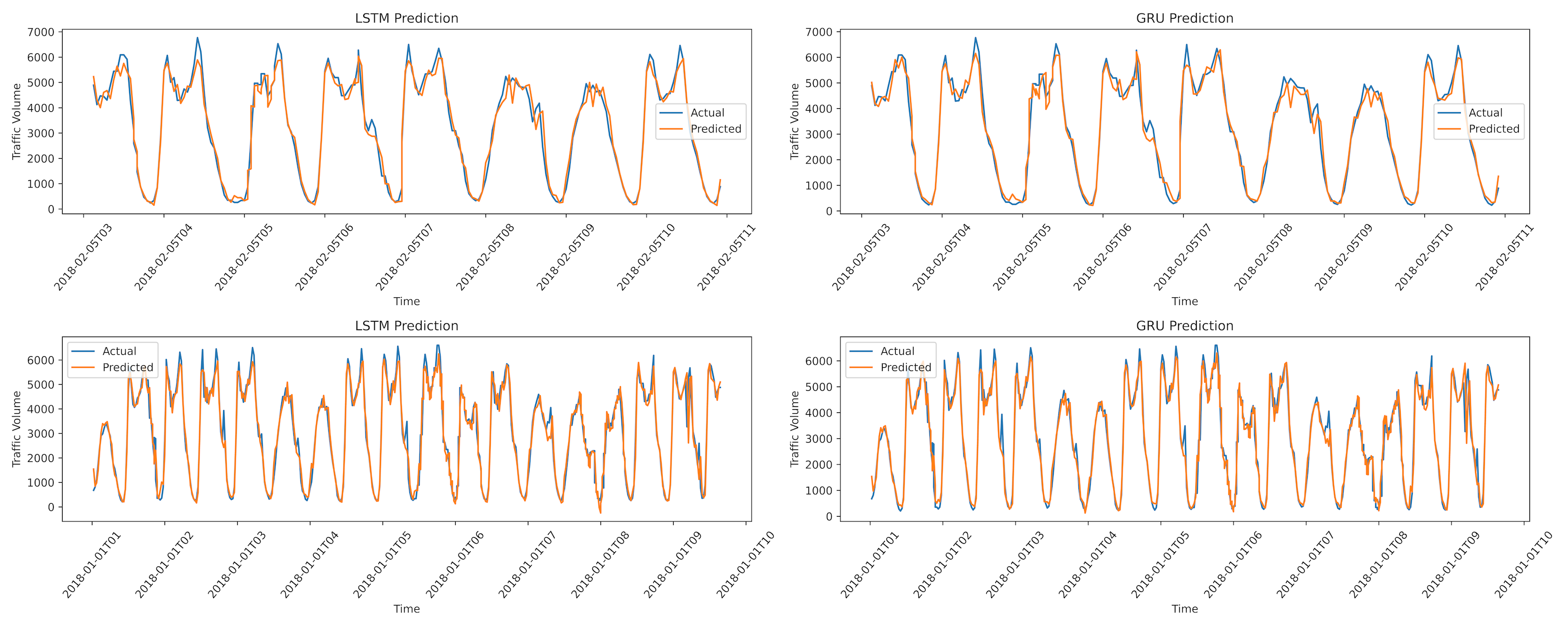}
    \caption{Traffic Volume Prediction at every hour using 6 hours of past observations as input. The models are able to produce very close to the actual traffic volume.}
    \label{fig:prediction_1}
\end{figure*}
\begin{figure*}[ht!]
    \centering
    \includegraphics[width=\textwidth]{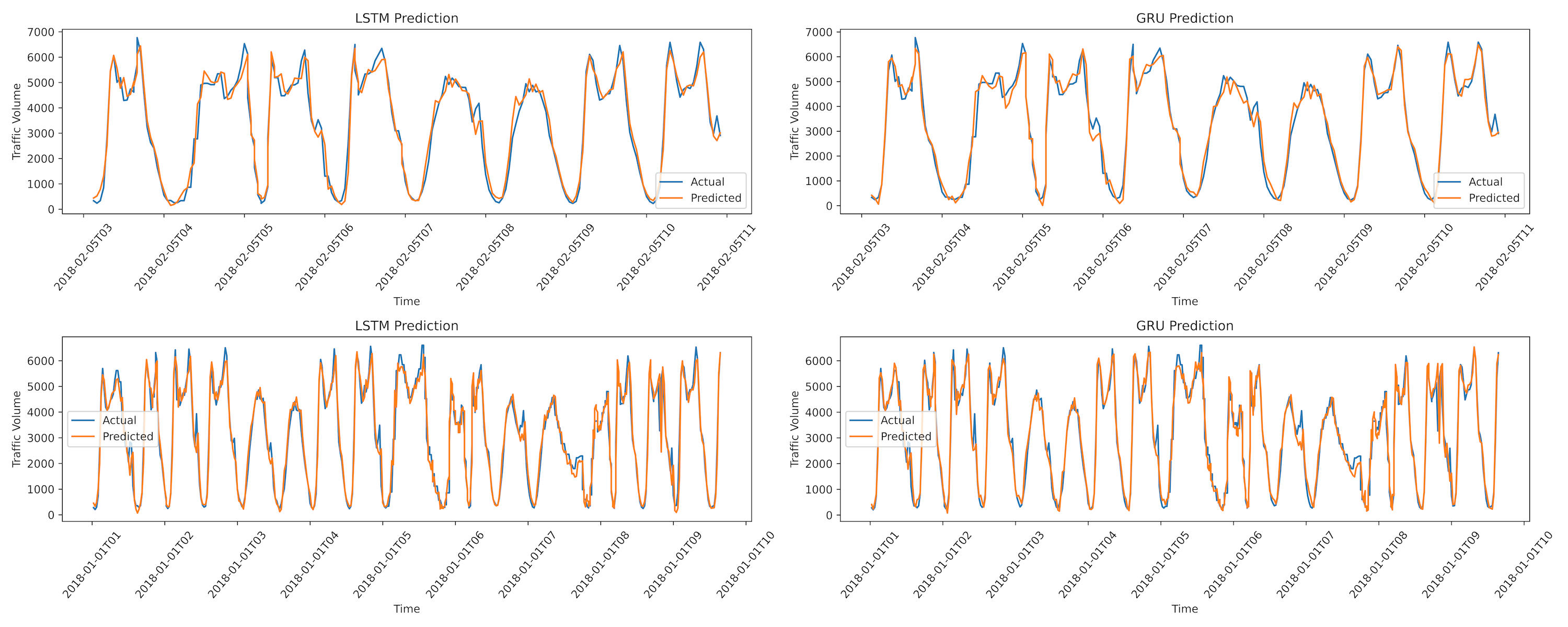}
    \caption{Traffic Volume Prediction at every hour using 24 hours of past observations as input. The models are able to produce very close to the actual traffic volume.}
    \label{fig:prediction_2}
\end{figure*}
Fig.\ref{fig:prediction_1} and Fig.\ref{fig:prediction_2} depict the predicted traffic volume results at various times. Here, the blue line indicates the actual traffic volume at that particular hour and the orange line represents the predicted traffic volume for that particular hour using the proposed models. The results clearly indicate that the LSTM and GRU can potentially provide accurate traffic volume as close to the actual values as possible. However, GRU outperforms LSTM when the past observation sequences are small, and LSTM performs better when the dataset is complex and requires the use of very long sequences to predict future traffic volume.
\begin{figure*}[ht!]
    \centering
    \includegraphics[width=\textwidth, height=0.3\linewidth]{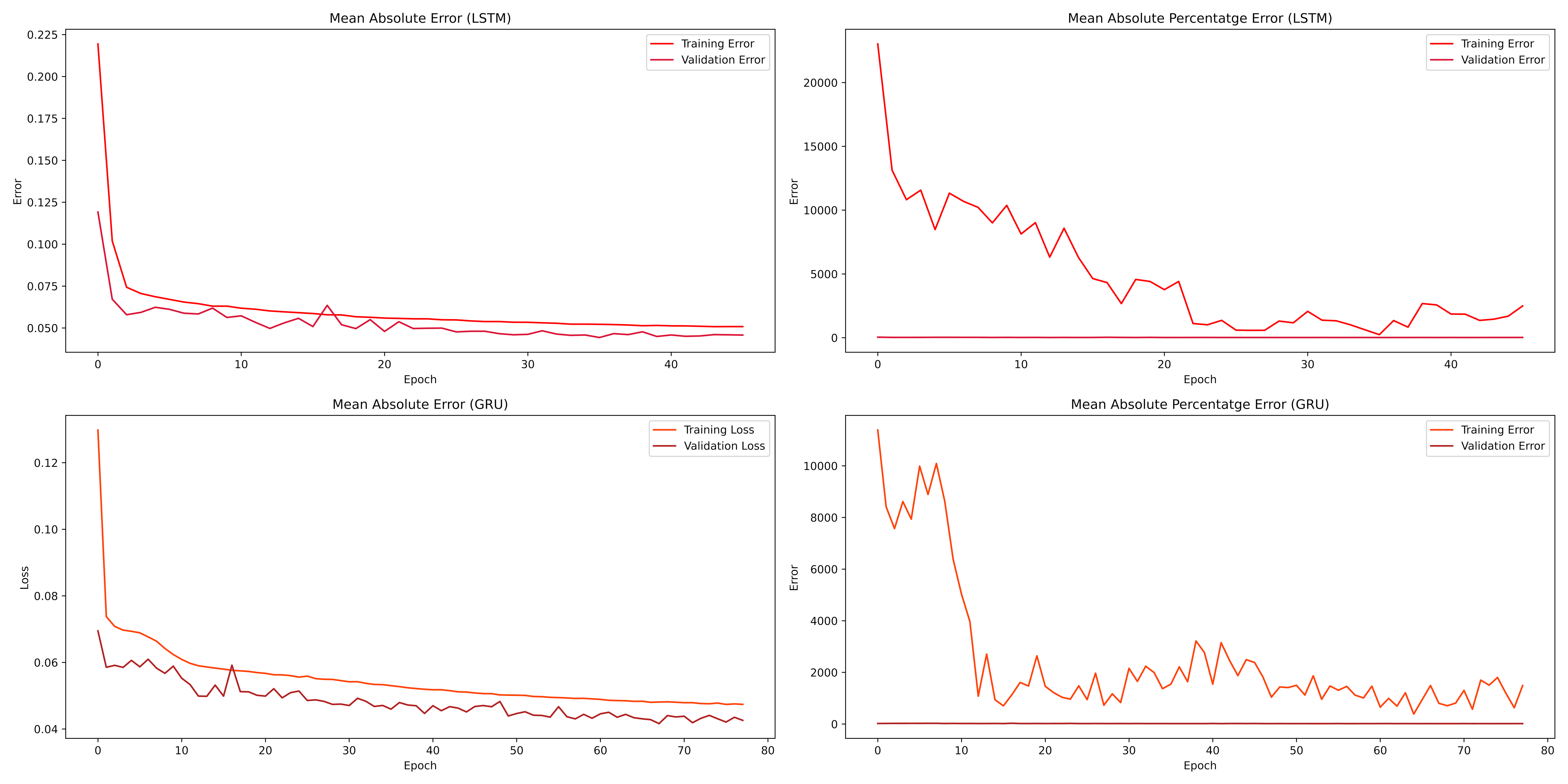}
    \caption{Training convergences for LSTM and GRU}
    \label{fig:training_convergence}
\end{figure*}
Fig.\ref{fig:training_convergence} shows the convergence of the LSTM and GRU models. Both models take more training time to get convergent when the number of features is less. However, GRU takes more time compared to LSTM. The reason may be that GRU does not use any memory units to control the flow of information.
\section{Discussion}
In this section, I will discuss challenges, limitations, computational efficiency, and potentially other methods for predicting traffic volume. The \href{https://github.com/lokesh-c-das/TrafficVolumePrediction}{Traffic Volume Prediction} repository contains the implementation, dataset, and trained models. Instructions on how to run can be found in the README.md file.
\subsection{Challenges and Limitations}
Preparing the dataset to feed into the LSTM and GRU networks was one of the more challenging tasks as it added another dimension (e.g., time) to the input shape. Moreover, to find a proper set of hyperparameters, the network required a significant amount of time, which was very challenging given the limited amount of computational resources. Finally, we have used MinMaxScaler to normalize the dataset. Whenever we compared the results with ground truth, we had to convert them back to their original format, which also requires a lot of work.
\subsection{Computational Time and Memory}
LSTM and GRU are computationally expensive and require a lot of memory, as they use memory to store historical observations internally. To finish a single training epoch given the above network structures and hyperparameters settings, the LSTM and GRU took approximately $25\sim 28s$.

\section{Conclusion}
In this project, we develop memory-based deep recurrent neural network models, namely LSTM and GRU, to predict traffic volume. The models are evaluated on the widely used Metro Interstate Traffic Volume dataset. The experimental results demonstrate the effectiveness of the proposed LSTM and GRU models. Recently, many researchers used graph-based attention modules to predict time series data. Graph-based multi-attention networks could be applied to predict traffic volume as well. To obtain better predictions, however, complete hyperparameter tuning and extensive experiments are required, which are our future works.
\bibliographystyle{IEEEtran}
\bibliography{mybibfile}

\end{document}